\documentclass[runningheads]{llncs}

\usepackage{eccv}

\usepackage{eccvabbrv}

\usepackage{graphicx}
\usepackage{booktabs}

\usepackage[accsupp]{axessibility}  

\usepackage{hyperref}

\usepackage{orcidlink}
\usepackage{textcomp}
\usepackage{bm}
\usepackage{threeparttable}
\usepackage{multirow}
\usepackage{algpseudocode}
\usepackage{algorithm}
\newcommand{\FLIP}{\protect\reflectbox{F}LIP\xspace}

\begin{document}
\begin{sloppypar}

\title{Multiscale Sliced Wasserstein Distances as Perceptual Color Difference Measures}

\titlerunning{MS-SWDs as Perceptual CD Measures}

\author{Jiaqi He\inst{1,2}\orcidlink{0000-0001-9768-2156} \and
Zhihua Wang\inst{3}\orcidlink{0000-0002-4398-536X} \and
Leon Wang\inst{4}
\and
Tsein-I Liu\inst{4}
\and
Yuming Fang\inst{5}\orcidlink{0000-0002-6946-3586}
\and
Qilin Sun\inst{6}\thanks{Corresponding author.}
\and
Kede Ma\inst{1,2}\orcidlink{0000-0001-8608-1128}}

\authorrunning{He~\etal}

\institute{Department of Computer Science, City University of Hong Kong 
\and Shenzhen Research Institute, City University of Hong Kong 
\and Department of Engineering, Shenzhen MSU-BIT University
\and Guangdong OPPO Mobile Telecommunications Corp., Ltd.
\and School of Information Management, Jiangxi University of Finance and Economics
\and School of Data Science, The Chinese University of Hong Kong (Shenzhen)\\
\email{jqhe00@mail.ustc.edu.cn} \quad
\email{zhihua.wang@my.cityu.edu.hk} \\
\email{\{leon.wang, simon\}@oppo.com} \quad
\email{fa0001ng@e.ntu.edu.sg} \\
\email{sunqilin@cuhk.edu.cn} \quad
\email{kede.ma@cityu.edu.hk}
}

\maketitle

\begin{abstract}
  Contemporary color difference (CD) measures for photographic images typically operate by comparing \textit{co-located} pixels, patches in a ``perceptually uniform'' color space, or features in a learned latent space. Consequently, these measures inadequately capture the human color perception of misaligned image pairs, which are prevalent in digital photography (\eg, the same scene captured by different smartphones). In this paper, we describe a perceptual CD measure based on the multiscale sliced Wasserstein distance, which facilitates efficient comparisons between \textit{non-local} patches of similar color and structure. This aligns with the modern understanding of color perception, where color and structure are inextricably interdependent as a unitary process of perceptual organization. Meanwhile, our method is easy to implement and training-free. Experimental results indicate that our CD measure performs favorably in assessing CDs in photographic images, and consistently surpasses competing models in the presence of image misalignment. Additionally, we empirically verify that our measure functions as a metric in the mathematical sense, and show its promise as a loss function for image and video color transfer tasks. The code is available at \url{https://github.com/real-hjq/MS-SWD}.

\end{abstract}

\section{Introduction}
\label{sec:intro}

\begin{figure*}[t]
    \centering
    \includegraphics[width=0.94\textwidth]{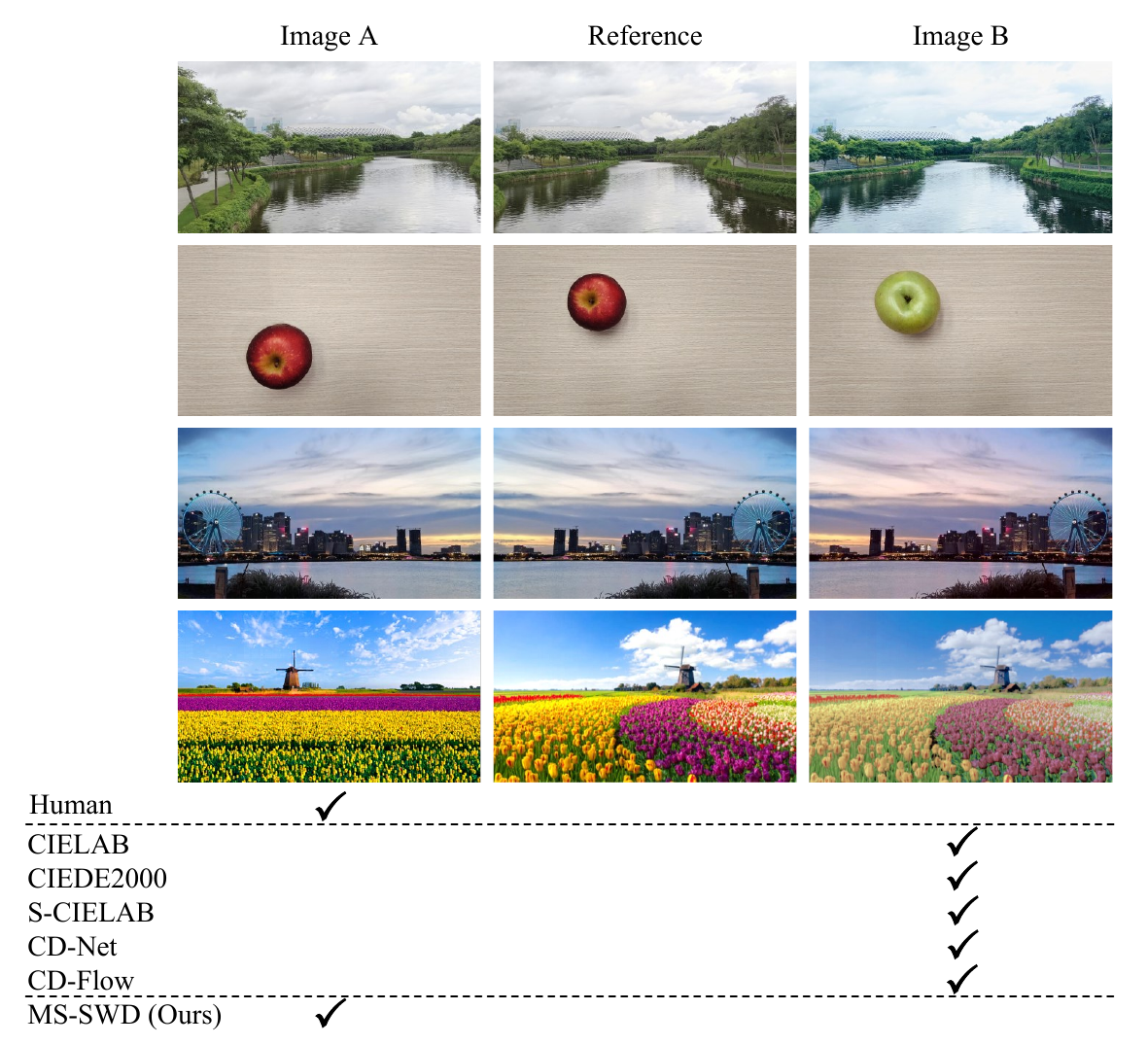}
    \caption{\textit{Which image is closer to the reference in terms of color appearance}?  Contemporary CD measures that seek \textit{co-located} comparisons often fail to explain human judgments. The proposed MS-SWD measure based on the multiscale sliced Wasserstein distance aligns with human color perception in these four challenging cases of image misalignment: global motion due to camera movement (first row), local motion due to object displacement (second row), horizontal flipping (third row), and similar natural scenes from different viewpoints (last row).}
   \label{fig:imgmisaligment}
\end{figure*}

Measuring perceptual color differences (CDs) in photographic images is a prerequisite in many image processing and computer vision tasks~\cite{sharma2017digital}. The predominant and scientifically well-founded theme is the pursuit of a perceptually uniform color space. Within such a space, numerical distances of two color points correspond directly to perceptual differences, regardless of their positions within the color spectrum. CIELAB and CIELUV, introduced by the Commission Internationale de l'{\'E}clairage (CIE) in 1976, represent two of the pioneering perceptually uniform color spaces~\cite{mahy1994cielab}. CD metrics (\eg, CIELAB $\Delta E^\star_{ab}$) derived from these color spaces have been rapidly adopted in various industrial sectors. However, subsequent analysis revealed that these color spaces are insufficient for accurately quantifying small to medium CDs~\cite{fairchild2013color}. In response, more sophisticated metrics (\eg, CIEDE2000~\cite{luo2001ciede2000}) were introduced to address various aspects of perceptual non-uniformity, whose rectified parameters were determined by fitting chromaticity discrimination (\ie, MacAdam) ellipses~\cite{luo2001ciede2000} obtained from subjective experiments.

Traditional CD metrics have demonstrated efficacy in predicting perceived differences between uniformly colored patches~\cite{moroney2002ciecam02}. A straightforward adaptation for assessing photographic images of natural scenes involves averaging the CDs between \textit{co-located} pixels~\cite{chen2023learning}. However, this na\"{i}ve extension shows a marginal correlation with human color perception, particularly when various sources of image misalignment are present (see Fig.~\ref{fig:imgmisaligment}).

Over the past decades, an extensive body of psychophysical and perceptual studies~\cite{lennie1999color, shevell2008color, ben2004hue, abertazzi2011perception} has provided a more compelling understanding of color perception: \textit{color, structure, and motion are inextricably interdependent as a unitary process of perceptual organization}~\cite{shapley2011color, kanizsa1979organization}. Drawing inspiration from these scientific insights, researchers have started to incorporate spatial modeling as a crucial component of  CD measures~\cite{zhang1997scielab, hong2006new, wang2004image, jaramillo2019evaluation, wang2022cdnet, chen2023learning, wang2024cdinet}. For instance, Zhang and Wandell~\cite{zhang1997scielab}  described a spatial extension of CIELAB $\Delta E^*_{ab}$ by applying lowpass filtering in an opponent color space as a preprocessing step. Wang \etal~\cite{wang2022cdnet} adopted a deep learning approach, training a lightweight neural network for ``color space transform'', followed by a learned Mahalanobis metric for distance calculation. Again, these models are designed to compare \textit{co-located} patches or features, making them susceptible to image misalignment (see Fig.~\ref{fig:imgmisaligment}).

In this paper, we introduce a perceptual CD measure that facilitates efficient comparisons between \textit{non-local} patches of similar color appearance and structural information. Our measure is primarily inspired by the seminal work of Elnekave and Weiss~\cite{elnekave2022generating}, who generated natural images by direct patch distribution matching. In a similar spirit, we compute the perceptual CD between two photographic images as the statistical distance of their patch distributions across multiple scales. To compare two images, we start by building two Gaussian pyramids in a perceptually more uniform CLELAB color space. Next, we opt for the sliced Wasserstein distance (SWD)~\cite{rabin2012wasserstein} to calculate the CD between the images at each scale. Finally, we average these CD values across all scales to obtain the overall CD estimate. The resulting measure, the multiscale SWD (MS-SWD), is conceptually simple and respects the modern view that color and structure interact inextricably in visual cortical processing. Meanwhile, MS-SWD is easy to implement and training-free.

We validate the proposed MS-SWD on the large-scale SPCD dataset~\cite{wang2022cdnet}. Remarkably, even without training, MS-SWD excels in evaluating CDs in photographic images, especially when there is large image misalignment. Additionally, we empirically show that MS-SWD behaves as a metric in the mathematical sense, and serves as a valid loss function for perceptual optimization in image and video color transfer tasks.

\setcounter{footnote}{0}
\section{Related Work}
\label{sec:relatedwork}
In this section, we present an overview of two areas of research closely related to our work: CD measures and patch matching methods in computer vision.
\subsection{CD Measures}
The development of CD measures has a rich history. In 1976,  CIE recommended the CIELAB color space~\cite{robertson1977cielab}, in which the Euclidean distance, $\Delta E^*_{ab}$, has been widely accepted as the de facto CD metric. Shortly after its introduction, researchers realized that CIELAB is not perfectly perceptually uniform. This led to the proposal of more sophisticated metrics such as CMC (l:c)~\cite{clarke1984cmc}, CIE94~\cite{mcdonald1995cie94}, and CIEDE2000~\cite{luo2001ciede2000}. These metrics generally assume a standard viewing environment, \eg, using the standard illuminant D65 and the $2^\circ$ standard observer, with reference to a white background, and do not explicit account for varying viewing conditions and ambient environments. To address this, metrics like CIECAM02~\cite{luo2013cam02} and CIECAM16~\cite{li2016cam16} were developed to predict changes in color appearance under varying viewing conditions. CIELAB-based methods are best suited for matching uniformly colored patches.

When assessing CDs in photographic images of natural scenes, humans tend to compare similar regions, co-located or not, in terms of color appearance and structural information within a broader spatial context~\cite{shapley2011color, kanizsa1979organization}. Zhang and Wandell~\cite{zhang1997scielab} made one of the first attempts to extend CIELAB to S-CIELAB by incorporating spatial lowpass filtering as front-end preprocessing. Choudhury \etal~\cite{choudhury2021image} designed preprocessing filters based on the contrast sensitivity functions (CSFs) of the human eye. Hong \etal~\cite{hong2006new} computed the weighted sum of pixel-wise CDs, prioritizing spatially homogeneous regions that cover large areas or have large predicted CDs. Similarly, Ortiz-Jaramillo~\etal~\cite{jaramillo2019evaluation} weighted patch-wise CDs using an image segmentation map computed from local binary patterns. The effectiveness of these spatially extended CD measures has been demonstrated only on small-scale private datasets with a few hand-picked images. Close to ours, Lee~\etal~\cite{lee2005evaluation} enabled \textit{non-local} CD assessment by histogram intersection\footnote{Histogram intersection measures the similarity between two normalized histograms by summing the minimum values of corresponding bins.}, which, however, completely throws away spatial information that is crucial for human color perception. Wang \etal~\cite{wang2022cdnet} demonstrated on the large-scale SPCD dataset that these simple spatial extensions may not yield noticeable performance improvements. As a result, they took a deep learning approach, and trained CD-Net~\cite{wang2022cdnet} and CD-Flow~\cite{chen2023learning} directly on SPCD. Inspired by~\cite{elnekave2022generating}, we tackle CD assessment of photographic images through multiscale patch distribution matching. MS-SWD enables efficient \textit{non-local} patch comparisons without using any specialized training.

\subsection{Patch Matching Methods} 
Patch matching is a fundamental technique in computer vision with diverse applications such as image denoising, image stitching, texture synthesis, image and video completion, 3D reconstruction, and object recognition. In patch matching, the search for patch nearest neighbors is often computationally intensive due to the need to explicitly establish bidirectional mappings~\cite{simakov2008summarizing,kolkin2019style,barnes2009patchmatch}.  In recent years, generative adversarial networks (GANs) and their derivatives have largely overtaken traditional patch matching methods. SinGAN~\cite{shaham2019singan} and InGAN~\cite{shocher2019ingan} are representative examples that indirectly match patch distributions of two images by training patch-based discriminators. To seek a direct (patch) distribution matching without involving time-consuming training, SWD has been explored in various image generation tasks,  in the raw pixel domain~\cite{elnekave2022generating,karras2018progressive,deshpande2018generative}, wavelet domain~\cite{rabin2012wasserstein}, and VGG feature domain~\cite{santos2019learning}. Our MS-SWD measure draws significant inspiration from~\cite{elnekave2022generating} but for a different purpose (\ie, CD assessment) with a different motivation (\ie, non-local patch comparison).

\section{MS-SWDs as Perceptual CD Measures}
\label{sec:method}
\begin{figure*}[t]
  \centering
  \includegraphics[width=1.0\linewidth]{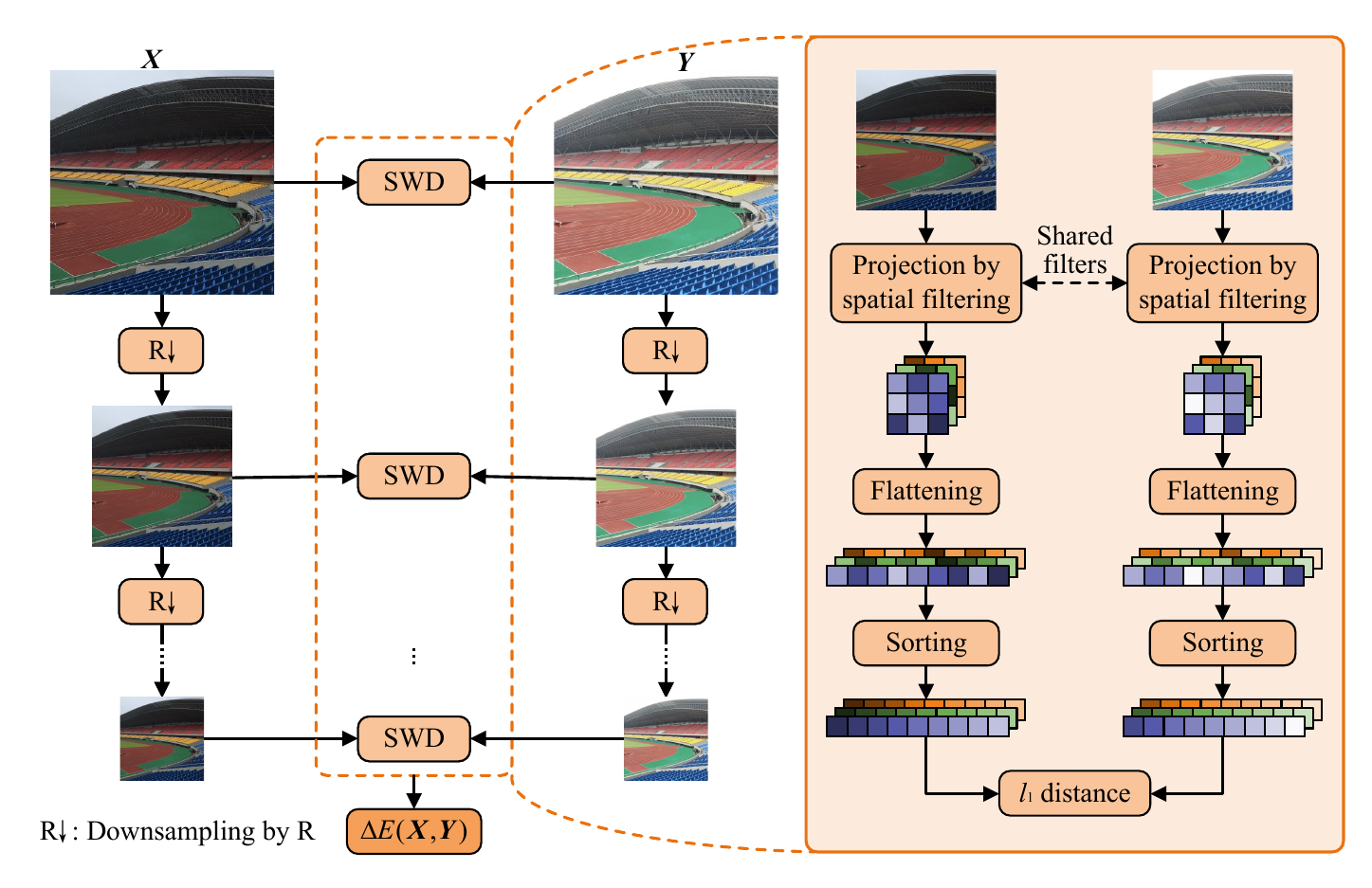}
  \caption{System diagram of the proposed MS-SWD for perceptual CD assessment.}
  \label{fig:cd_metirc}
\end{figure*}

In this section, we first introduce the necessary preliminaries - SWD, and then present in detail our MS-SWD measure for perceptual CD assessment. Fig.~\ref{fig:cd_metirc} shows the system diagram of MS-SWD.

\subsection{SWD}
Among various statistical distances between two probability distributions, the Wasserstein distance enjoys several advantages, including 1) intuitive interpretation (as the minimum ``cost'' of transforming one distribution into another), 2) sensitivity to distribution shape (by computing the actual geometric distances between points in the distributions), 3) robustness to support differences (even when the supports of the two distributions do not overlap), and 4) smooth gradients for optimization~\cite{arjovsky2017wasserstein}. The  $1$-Wasserstein distance (also known as the earth mover's distance) between two probability distributions $\mu$ and $\nu$ is defined as 
\begin{equation}
\mathrm{WD}(\mu, \nu)= \inf _{\gamma \in \Gamma(\mu, \nu)} \mathbb{E}_{(\bm x, \bm y) \sim \gamma}\Vert \bm x- \bm y\Vert_1,
\end{equation}
where $\Gamma(\mu, \nu)$ denotes the set of all joint distributions (couplings) $\gamma$ whose marginals are $\mu$ and $\nu$. The Wasserstein distance is notoriously challenging to implement due to its high computational complexity, especially when working with empirical distributions represented by high-dimensional samples\footnote{This corresponds to solving a large-scale linear programming problem, which is painfully slow.}. To reduce the computational complexity and improve the scalability and robustness to high dimensions, Rabin~\etal~\cite{rabin2012wasserstein} introduced SWD by projecting the high-dimensional data onto a lower-dimensional subspace and then calculating the Wasserstein distance therein. When the projected space is one-dimensional, SWD can be mathematically expressed as 
\begin{equation}
\mathrm{SWD}(\bm{U}, \bm{V})=E_{\bm{w}\sim\mathcal{U}(\mathbb{S}^{N-1})}\mathrm{WD}\left(\bm{U}\bm{w}, \bm{V}\bm{w}\right),
\label{eq:swd_expect}
\end{equation}
where $\bm U, \bm V\in\mathbb{R}^{M\times N}$, $M$ is the number of samples to represent the empirical distributions, and $N$ is the sample dimension. $\mathbb{S}^{N-1}:=\left\{\bm{w} \in \mathbb{R}^{N\times 1} \mid\Vert\bm{w}\Vert_2^2=1\right\}$ for any $N \geq 2$ is the unit hyper-sphere, $\mathcal{U}\left(\mathbb{S}^{N-1}\right)$ is the uniform distribution defined over $\mathbb{S}^{N-1}$, and $\mathbb{E}_{\bm{w}}$ is the expectation over the random unit vector $\bm{w}$. In Eq.~\eqref{eq:swd_expect}, the one-dimensional Wasserstein distance can be efficiently calculated by \textit{sorting} the projected samples and computing the $\ell_1$-distance between the sorted samples~\cite{shen1983generalized}. SWD typically reduces the computational complexity from $\mathcal{O}(M^{2.5})$~\cite{pitie2005n} to $\mathcal{O}(M \log M)$.

\subsection{MS-SWD for CD Assessment}

\begin{algorithm}[t]
\caption{MS-SWD for Perceptual CD Assessment}\label{algorithm}
\begin{algorithmic}[1]
  \State \textbf{Input}: A pair of photographic images that are possibly misaligned, ($\bm{X}, \bm{Y}$), the number of scales, $K$, and the number of random projections, $P$
  \State \textbf{Output}: Predicted CD, $\Delta E(\bm{X}, \bm{Y})$
  \State Build Gaussian pyramids $\{\bm{X}^{(i)}\}^{K}_{i=1}$ and $\{\bm{Y}^{(i)}\}^K_{i=1}$, where $\bm{X}^{(1)} = \bm{X}$ and  $\bm{Y}^{(1)} = \bm{Y}$
  \State Convert $\{\bm{X}^{(i)}\}^{K}_{i=1}$ and $\{\bm{Y}^{(i)}\}^{K}_{i=1}$ from the sRGB to CIELAB color space
  \State $\Delta E \leftarrow 0$
  \For{$i \leftarrow 1$ \textbf{to} $K$}
    \For{$j \leftarrow 1$ \textbf{to} $P$} 
    \State $\bm{w}\sim\mathcal{U}(\mathbb{S}^{N\times 3-1})$
    \State $\bm{w} \leftarrow \operatorname{unflat}(\bm{w})$
    \Comment{``unflat()'' converts a vector into a tensor}
    \State $\bm{x} \leftarrow \operatorname{flat}(\operatorname{Conv2d}(\bm{X}^{(i)}, \bm{w}, $ \textquotesingle reflect\textquotesingle)) 
    \Comment{``flat()'' is the inverse of ``unflat()''}
    \State $\bm{y} \leftarrow \operatorname{flat}(\operatorname{Conv2d}(\bm{Y}^{(i)}, \bm{w}, $ \textquotesingle reflect\textquotesingle))  
    \State $\Delta E \leftarrow \Delta E + \frac{1}{M}\Vert\operatorname{sort}(\bm{x})- \operatorname{sort}(\bm{y})\Vert_1$\label{step}
    \EndFor
  \EndFor
\State $\Delta E(\bm{X}, \bm{Y}) \leftarrow \frac{1}{KP}\Delta E$
\end{algorithmic}
\end{algorithm}

 Let $\bm X \in \mathbb{R}^{H\times W\times 3}$ and $\bm Y \in \mathbb{R}^{H\times W\times 3}$ be two photographic images that are possibly misaligned, where $H$ and $W$ are the image height and width, respectively. We first construct two Gaussian pyramids, $\{\bm{X}^{(i)}\}_{i=1}^{K}$ and $\{\bm{Y}^{(i)}\}_{i=1}^{K} $ by iteratively applying a Gaussian filter and downsampling the filtered image by a factor of $R$, where $\bm{X}^{(i)},\bm{Y}^{(i)}\in\mathbb{R}^{\lfloor H / 2^{i-1}\rfloor \times \lfloor W / 2^{i-1}\rfloor \times 3 }$ and $K$ denotes the number of scales. We then represent $\bm {X}^{(i)}$ and $\bm {Y}^{(i)}$, for $1 \le i \le K$, in the CIELAB color space, where we observe significant performance gains over the sRGB color space. Although spatial pre-filtering of $\bm {X}^{(i)}$ and $\bm {Y}^{(i)}$ based on CSFs~\cite{zhang1997scielab,choudhury2021image} can also be applied, it does not yield noticeable improvements and is therefore excluded from our current implementation.

For ease of mathematical description, we also use an alternative notation for $\bm{X}^{(i)}$, denoted as $\bm{X}_\mathrm{col}^{(i)}\in \mathbb{R}^{M\times (N\times 3)}$, in which we rearrange the $M$ overlapping image patches of size $\sqrt{N}\times \sqrt{N}\times 3$ into columns. The transformation from $\bm {X}^{(i)}$ to $\bm{X}_\mathrm{col}^{(i)}$ can be efficiently achieved using the \texttt{img2col()} operator, a common tool in image processing for implementing convolutions.

\begin{figure}[t]
    \centering
    \includegraphics[width=0.99\linewidth]{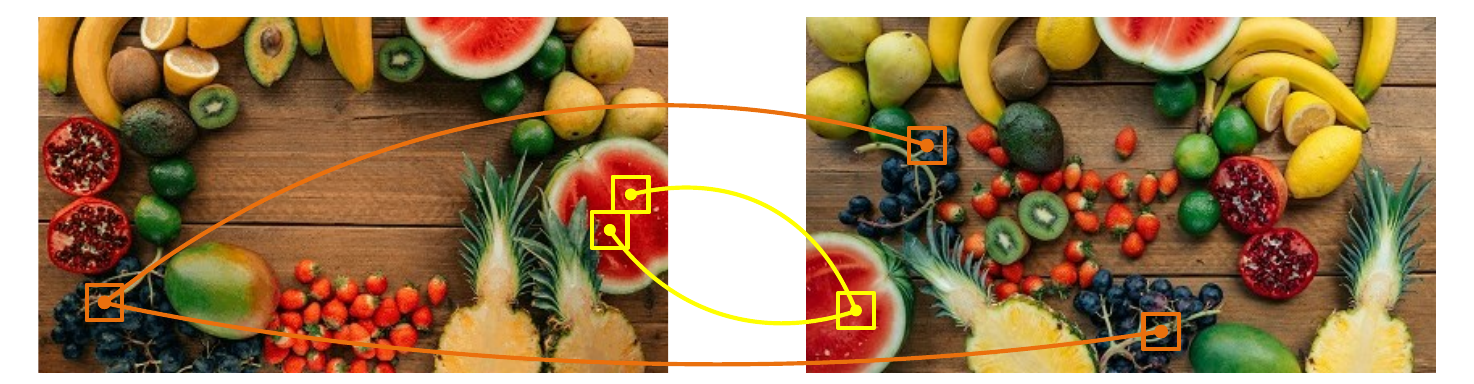}\label{fig:comparison}
    \caption{The \texttt{sort()} operator in MS-SWD enables efficient comparisons of non-local patches with similar color appearance and structural information. Each curve represents a different random projection; the patches at the two ends of the curve share the same rank (\ie, correspondence) after sorting, thus subject to CD calculation.}
   \label{fig:non-local patches}
\end{figure}

After generating the Gaussian pyramids $\{\bm{X}^{(i)}_\mathrm{col}\}_{i=1}^{K}$ and $\{\bm{Y}^{(i)}_\mathrm{col}\}_{i=1}^{K}$ in CIELAB, we calculate the predicted CD, $\Delta E(\bm X, \bm Y)$, between the two images $\bm X$ and $\bm Y$ by averaging SWD from Eq.~\eqref{eq:swd_expect} across all scales:
\begin{equation}
  \Delta E(\bm{X}, \bm{Y})= \frac{1}{K}\sum_{i=1}^K\mathrm{SWD}\left(\bm{X}_\mathrm{col}^{(i)}, \bm{Y}_\mathrm{col}^{(i)}\right)=\frac{1}{KP}\sum_{i=1}^K{\sum_{j=1}^P{\mathrm{WD}\left(\bm{X}_\mathrm{col}^{(i)}\bm{w}^{(j)}, \bm{Y}_\mathrm{col}^{(i)}\bm{w}^{(j)}\right)}},
  \label{eq:SWD_pyramid}
\end{equation}
where the expectation in Eq.~\eqref{eq:swd_expect} is approximated by the average over a set of $P$ random unit projections $\{\bm w^{(j)}\}_{j=1}^P$. It is important to note that  the matrix multiplication, $\bm{X}_\mathrm{col}^{(i)}\bm{w}^{(j)}$ (and $\bm{Y}_\mathrm{col}^{(i)}\bm{w}^{(j)}$) can be implemented by a single convolution. The computational procedure of MS-SWD for perceptual CD assessment is given in Algorithm~\ref{algorithm}, in which we particularly emphasize the importance of the \texttt{sort()} operator in Step~\ref{step}. Provided that the two images differ primarily in color appearance, patches of similar color and structure, whether co-located or not, are likely to have similar ranks (\ie, correspondences) in the projected space after sorting, thus subject to CD calculation. This 1) facilitates efficient non-local patch comparisons without the need to compute patch nearest neighbors~\cite{elnekave2022generating}, and meanwhile 2) respects the modern view of human color perception~\cite{abertazzi2011perception} that color and structure are inextricably interdependent as a unitary process of perceptual organization~\cite{shapley2011color,kanizsa1979organization}.

\begin{figure}[t]
    \centering
    \subfloat[Reference $\bm X$]{\includegraphics[width=0.24\linewidth]{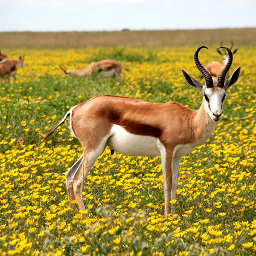}\label{fig:img_recovery_a}}\hfill
    \subfloat[Initial $\bm{Y}_\mathrm{init}$]{\includegraphics[width=0.24\linewidth]{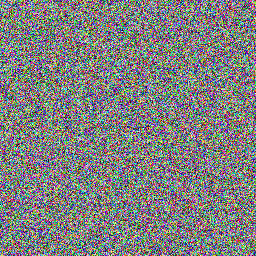}\label{fig:img_recovery_b}}\hfill
    \subfloat[1 scale]{\includegraphics[width=0.24\linewidth]{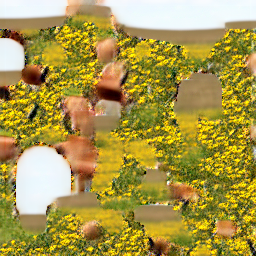}\label{fig:img_recovery_c}}\hfill
    \subfloat[2 scales]{\includegraphics[width=0.24\linewidth]{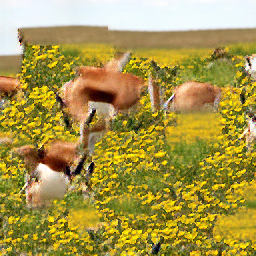}\label{fig:img_recovery_d}}\\
    \subfloat[3 scales]{\includegraphics[width=0.24\linewidth]{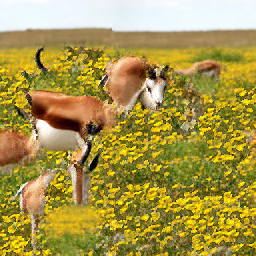}\label{fig:img_recovery_e}}\hfill
    \subfloat[4 scales]{\includegraphics[width=0.24\linewidth]{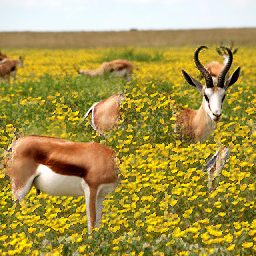}\label{fig:img_recovery_f}}\hfill
    \subfloat[5 scales]{\includegraphics[width=0.24\linewidth]{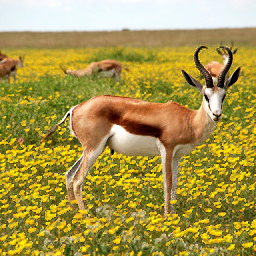}\label{fig:img_recovery_g}}\hfill
    \subfloat[6 scales]{\includegraphics[width=0.24\linewidth]{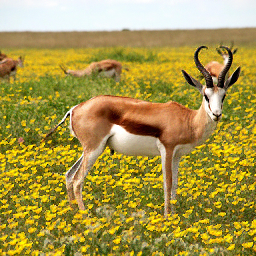}\label{fig:img_recovery_h}}
    \caption{Illustration of multiscale analysis in ensuring pixel-level image fidelity. Images (c)-(h) are generated by minimizing $\Delta E (\bm X, \bm Y)$ with respect to $\bm Y$ to match (a) the reference image $\bm X$, starting from (b) the initial Gaussian noise image $\bm{Y}_\mathrm{init}$ and for different values of $K$.}  \label{fig:img_recovery}
\end{figure}

Multiscale analysis is another crucial aspect of our approach, although, through our internal subjective testing, it seems that human color perception of photographic images remains fairly stable under varying viewing conditions related to image scale (\eg, display resolution and viewing distance). This is because matching the single-scale patch distribution is unlikely to guarantee image fidelity at the pixel level. As illustrated in Fig.~\ref{fig:img_recovery},  we begin with a Gaussian noise image $\bm{Y}_\mathrm{init}$ of the same size as the reference image $\bm X$, and iteratively refine $\bm{Y}_\mathrm{init}$ by minimizing Eq.~\eqref{eq:SWD_pyramid} of varying $K$ using gradient-based optimization. With a limited number of scales, the optimized image fails to recover the structural details of the reference, and exhibits perceptually annoying distortions (\eg, object discontinuity), despite the MS-SWD value being close to zero. Our empirical observations indicate that for a $256\times 256$ image, using five scales suffices to recover the reference image within the human perceptual threshold.

\vspace{-0.15cm}
\section{Experiments}
\label{sec:experiments}
In this section, we first compare the proposed MS-SWD with existing CD measures on SPCD~\cite{xu2022database, wang2022cdnet}. We then perform a series of ablation studies to validate the key design choices of MS-SWD. Finally, we explore the use of MS-SWD in guiding image and video color transfer.

\subsection{Main Experiments}
\textbf{SPCD Dataset}. 
We conduct the main experiments on SPCD~\cite{xu2022database, wang2022cdnet}, which is the largest image dataset currently available for CD assessment. SPCD comprises $30,000$ photographic image pairs that span diverse real-world picture-taking scenarios, featuring great variations in foreground elements, background complexity, lighting and weather conditions, and camera modes. Of these $30,000$ pairs, $10,005$ are non-perfectly aligned, captured by six flagship smartphones, while the remaining pairs are perfectly aligned with CDs induced through simulated color alterations.

\begin{table*}[t]
\scriptsize
    \caption{Performance evaluation of CD measures on the SPCD dataset. The top section lists standard CD formulae derived from uniformly colored patches. The second section contains CD measures adapted for photographic images. The third section includes general-purpose image quality models. The fourth section consists of just-noticeable difference measures. The top two methods are highlighted in boldface.}
    \label{tab:SPCD}
    \centering
    \begin{threeparttable} 
    \begin{tabular}{l|ccc|ccc|ccc}
        \toprule[1pt]
	\multirow{2}*{Method}  & \multicolumn{3}{c|}{Perfectly aligned pairs} & \multicolumn{3}{c|}{Non-perfectly aligned pairs} & \multicolumn{3}{c}{All}\\
        \cline{2-10}
        & STRESS$\downarrow$ & PLCC$\uparrow$ & SRCC$\uparrow$ & STRESS$\downarrow$ & PLCC$\uparrow$ & SRCC$\uparrow$ & STRESS$\downarrow$ & PLCC$\uparrow$ & SRCC$\uparrow$\\
        \hline
 	CIELAB~\cite{robertson1977cielab} &$31.280$ & $0.790$ & $0.774$ & $30.009$ & $0.683$ & $0.577$  & $31.952$ & $0.714$ & $0.665$   \\
 	CIE94~\cite{mcdonald1995cie94}  &$34.643$ & $0.786$ & $0.772$ & $30.147$ &  $0.692$ & $0.572$  & $34.305$ & $0.709$ & $0.654$   \\
        CIEDE2000~\cite{luo2001ciede2000}  &$29.862$ & $\textbf{0.827}$ & $\textbf{0.821}$ & $30.650$ &  $0.653$ & $0.561$  & $31.431$ & $0.725$ & $0.685$   \\
        CIECAM02~\cite{luo2013cam02}  &$\textbf{24.779}$ & $0.823$ & $0.820$ & $29.339$ &  $0.679$ & $0.612$  & $\textbf{27.151}$ & $\textbf{0.748}$ & $0.725$   \\
        CIECAM16~\cite{li2016cam16}  &$\textbf{23.901}$ & $0.818$ & $0.820$ & $29.934$ & $0.661$ & $0.600$  & $\textbf{26.817}$ & $0.743$ & $\textbf{0.726}$   \\
 	\hline
 	S-CIELAB~\cite{zhang1997scielab} &$29.977$ & $0.824$ & $0.819$ & $32.057$ & $0.627$ & $0.522$  & $32.760$ & $0.699$ & $0.657$   \\   
        Lee05~\cite{lee2005evaluation} & $58.652$ & $0.728$ & $0.735$ & $56.515$ & $0.636$ & $0.637$  & $58.031$ & $0.697$ & $0.710$   \\
        Hong06~\cite{hong2006new} &$60.361$ & $0.732$ & $0.811$ & $57.466$ &  $0.538$ & $0.462$ & $61.242$ & $0.609$ & $0.634$   \\
 	Ouni08~\cite{ouni2008new} &$29.864$ & $\textbf{0.826}$ & $\textbf{0.821} $ & $30.657$ &  $0.653$ & $0.561$  & $31.435$ & $0.722$ & $0.685$   \\		
 	\hline
 	PieAPP~\cite{prashnani2018pieapp} &$41.550$ & $0.502$ & $0.511$ & $39.619$ &  $0.483$ & $0.410$  & $41.896$ & $0.467$ & $0.451$   \\
 	LPIPS~\cite{zhang2018lpips} &$40.972$ & $0.767$ & $0.766$ & $46.402$ &  $0.272$ & $0.237$  & $64.407$ & $0.448$ & $0.396$   \\  
   	\FLIP~\cite{andersson2020flip} &$29.368$ & $0.743$ & $0.714$ & $\textbf{27.559}$ & $\textbf{0.730}$ & $\textbf{0.638}$  & $29.197$ & $0.715$ & $0.663$   \\
 	DISTS~\cite{ding2020image} &$33.417$ & $0.725$ & $0.722$ & $33.244$ &  $0.571$ & $0.495$  & $37.236$ & $0.582$ & $0.549$   \\
        A-DISTS~\cite{ding2021adists} & $38.190$ & $0.661$ & $0.663$ & $42.488$ & $0.387$ & $0.365$ & $51.360$& $0.424$& $0.384$\\
        ST-LPIPS~\cite{ghildyal2022stlpips} & $37.234$  &$0.810$ & $0.813$ & $43.912$ & $0.399$ & $0.362$ & $50.579$ & $0.535$ & $0.512$\\
        DeepWSD~\cite{liao2022DeepWSD} & $31.760$ & $0.539$ & $0.540$& $43.342$ & $0.055$ & $0.015$& $49.705$ & $0.136$ & $0.180$\\
 	\hline
 	Chou07~\cite{chou2007fidelity} &$52.463$ & $0.780$ & $0.793$ & $37.704$ & $0.645$ & $0.518$  & $49.581$ & $0.667$ & $0.615$   \\
 	Butteraugli~\cite{alakuijala2017guetzli} &$42.691$ & $0.615$ & $0.589$ & $48.764$ &  $0.205$ & $0.193$  & $54.801$ & $0.372$ & $0.354$  \\
        PIM-5~\cite{bhardwaj2020pim} & $58.737$ & $0.685$ & $0.695$  &  $48.454$  & $0.556$ & $0.482$  & $60.346$  &  $0.455$  & $0.480$\\
 	\hline
	MS-SWD (Ours) &  $34.040$  & $0.778$ & $0.755$ & $\textbf{28.363}$ & $\mathbf{0.841}$ & $\mathbf{0.805}$& $32.781$ & $\mathbf{0.794}$ & $\mathbf{0.772}$\\  
	\bottomrule[1pt]
    \end{tabular}
    \end{threeparttable}
\end{table*}

\noindent
\textbf{Implementation Details}. 
MS-SWD does not contain any trainable parameters; all its hyper-parameters are inherited directly from previous studies. These include the number of scales $K=5$, the downsampling factor $R=2$, and the filter size $\sqrt{N}=11$ with a stride of $1$ from the MS-SSIM paper~\cite{wang2003multiscale}, and the number of random unit projections $P=128$ from the GPDM paper~\cite{elnekave2022generating}. Throughout all experiments, we resize the images to $256\times 256$ for testing.

\noindent
\textbf{Evaluation Criteria}. We employ three evaluation criteria: standardized residual sum of squares (STRESS) \cite{garcia2007measurement}, Pearson linear correlation coefficient (PLCC), and Spearman's rank correlation coefficient (SRCC). 
STRESS assesses the prediction accuracy and statistical significance, and is defined as
\begin{equation}
  \text{STRESS} = 100\sqrt{\frac{\sum_{i=1}^{I}(\Delta E_i - F \Delta V_i)^2}{F^2 \sum_{i=1}^{I}(\Delta V_i)^2}},
  \label{eq:stess}
\end{equation}
where $I$ is the number of test pairs, and $F$ is the scale correction factor between predicted CDs, $\Delta E$ and ground-truth CDs, $\Delta V$:
\begin{equation}
  F=\frac{\sum_{i=1}^{I}(\Delta E_i)^2}{\sum_{i=1}^{I}\Delta E_i \Delta V_i}.
  \label{eq:factor}
\end{equation}
A smaller value of STRESS indicates a tighter fit. PLCC and SRCC measure the prediction linearity and monotonicity, respectively, with a larger value indicating better correlation. Before calculating PLCC, we linearize model predictions by fitting a four-parameter logistic function, as suggested in~\cite{video2000final}.

\noindent
\textbf{SPCD Results}. We compare MS-SWD against $19$ state-of-the-art methods, categorized as follows: 1) CD metrics derived from uniformly colored patches, including CIELAB \cite{robertson1977cielab}, CIE94 \cite{mcdonald1995cie94}, CIEDE2000 \cite{luo2001ciede2000}, CIECAM02 \cite{luo2013cam02}, and CIECAM16 \cite{li2016cam16}; 2) CD measures designed for photographic images, including S-CIELAB \cite{zhang1997scielab}, Lee05 \cite{lee2005evaluation}, Hong06 \cite{hong2006new}, and Ouni08 \cite{ouni2008new}; 3) general-purpose image quality models, including PieAPP \cite{prashnani2018pieapp}, LPIPS \cite{zhang2018lpips}, \FLIP \cite{andersson2020flip}, DISTS \cite{ding2020image},  A-DISTS \cite{ding2021adists}, ST-LPIPS \cite{ghildyal2022stlpips}, and DeepWSD \cite{liao2022DeepWSD}; 4) just-noticeable difference methods, including Chou07 \cite{chou2007fidelity}, Butteraugli \cite{alakuijala2017guetzli}, and PIM-5 \cite{bhardwaj2020pim}. We use the official implementations provided by the original authors for CIECAM02, CIECAM16, Butteraugli, PIM-5, and the seven general-purpose image quality models. For the remaining methods, we use the implementations provided by Jaramillo~\etal~\cite{jaramillo2019evaluation}.

From the results in \Cref{tab:SPCD}, we have several key observations. First, the majority of CD methods exhibit diminished performance for non-perfectly aligned pairs due to co-located comparisons, even when the misalignment is imperceptible to the human eye. Second, CD formulae recommended by CIE, along with their spatial extensions S-CIELAB and Ouni08, deliver outstanding correlation with human color perception, especially on perfectly aligned pairs. This provides a strong indication of the practical applicability of the CIELAB color space. Third, general-purpose image quality models and just-noticeable difference measures fail to accurately predict CDs in photographic images. Finally, the proposed MS-SWD significantly surpasses all competing methods on the non-perfectly aligned pairs, and achieves the overall best performance in terms of PLCC and SRCC without training on perceptual CD data. This highlights the importance of non-local patch comparisons in CD assessment. 

\begin{table*}[t]
\scriptsize
    \caption{Performance evaluation of CD measures on the augmented SPCD dataset by geometric transformations. ``Translation'' involves randomly shifting one image relative to the other by up to $5\%$ of pixels in both axes. ``Dilation'' refers to enlarging one image by a factor of $1.1$. ``Flipping'' means horizontally flipping one image.}
    \label{tab:SPCD_distortion}
    \centering
    \begin{tabular}{l|ccc|ccc|ccc}
    \toprule[1pt]
    \multirow{2}*{Method}  & \multicolumn{3}{c|}{Translation} & \multicolumn{3}{c|}{Dilation} & \multicolumn{3}{c}{Flipping}\\
    \cline{2-10}
    & STRESS$\downarrow$ & PLCC$\uparrow$ & SRCC$\uparrow$ & STRESS$\downarrow$ & PLCC$\uparrow$ & SRCC$\uparrow$ & STRESS$\downarrow$ & PLCC$\uparrow$ & SRCC$\uparrow$\\
    \hline    
    CIELAB~\cite{robertson1977cielab} &$38.271$ & $0.386$ & $0.304$ & $38.667$ & $0.361$ & $0.260$ & $42.956$ &  $0.168$ & $0.094$   \\
    CIE94~\cite{mcdonald1995cie94} &$37.271$ & $0.435$ & $0.318$  &  $37.539$ &  $0.419$ & $0.274$ & $41.715$ &  $0.210$ & $0.113$  \\
    CIEDE2000~\cite{luo2001ciede2000} &$38.365$ & $0.377$ & $0.284$ & $38.619$ & $0.362$ & $0.240$ & $42.770$ & $0.170$ & $0.079$   \\
    S-CIELAB~\cite{zhang1997scielab} &$39.048$ & $0.349$ & $0.253$ & $39.262$ & $0.332$ & $0.216$ & $42.960$ & $0.151$ & $0.065$    \\
    Lee05~\cite{lee2005evaluation} & $56.466$ & $0.632$ & $\textbf{0.633}$ & $56.529$ & $0.636$ & $\textbf{0.637}$ & $56.515$ & $\textbf{0.636}$ & $\textbf{0.637}$   \\
    Hong06~\cite{hong2006new} &$57.521$ & $0.297$ & $0.206$ & $55.609$ & $0.284$ & $0.177$ & $56.718$ &  $0.154$ & $0.098$ \\
    LPIPS~\cite{zhang2018lpips} &$45.853$ & $0.048$ & $0.018$  & $43.882$ & $0.083$ & $0.109$ & $43.545$ &  $0.074$ & $0.104$   \\
    DISTS~\cite{ding2020image} &$37.303$ & $0.362$ & $0.289$  & $37.519$ & $0.317$ & $0.252$ & $37.287$ &  $0.316$ & $0.233$  \\
    CD-Net~\cite{wang2022cdnet} & $29.737$ & $0.659$ & $0.567$ & $29.848$ & $0.656$ & $0.542$ & $39.325$ & $0.295$ & $0.221$ \\
    CD-Flow~\cite{chen2023learning} & $\textbf{29.188}$ & $\textbf{0.719}$  & $0.569$  & $\textbf{29.065}$  &  $\textbf{0.705}$  & $0.584$ & $\textbf{36.546}$  & $0.393$ & $0.263$  \\
    \hline
    MS-SWD (Ours) &  $\mathbf{28.353}$  & $\mathbf{0.836}$ & $\mathbf{0.798}$ & $\mathbf{28.144}$ & $\mathbf{0.833}$ & $\mathbf{0.793}$ & $\mathbf{26.132}$ & $\mathbf{0.836}$ & $\mathbf{0.788}$\\  
    \bottomrule[1pt]
    \end{tabular}
\end{table*}

\noindent
\textbf{Robustness Results to Geometric Transformations}. 
To further verify the robustness of MS-SWD to geometric transformations, we follow the experimental procedure in~\cite{chen2023learning}, and augment SPCD by 1) randomly shifting one image relative to the other by up to $5\%$ pixels in both axes, 2) enlarging one image by a factor of $1.1$, and 3) horizontally flipping one image (see the third row of Fig.~\ref{fig:imgmisaligment}). These transformations are applied to the non-perfectly aligned pairs in the SPCD dataset. From the results in \Cref{tab:SPCD_distortion}, we find that all competing methods experience a significant performance drop, except for Lee05 which benefits from non-local CD assessment by histogram intersection. Although designed to be aware of geometric transformations, DISTS, CD-Net, and CD-Flow can only handle mild transformations. In stark contrast, the proposed MS-SWD is exceptionally robust to geometric transformations, even in the challenging case of horizontal flipping.

\begin{figure}[t]
    \centering
    \subfloat[Image A]{\includegraphics[width=0.19\linewidth]{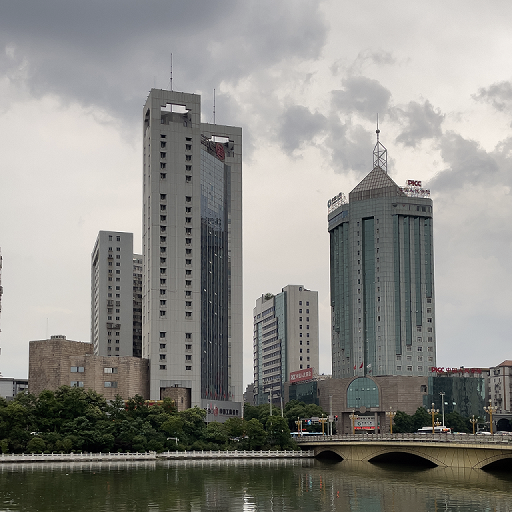}} \hfill
    \subfloat[CIELAB]{\includegraphics[width=0.19\linewidth]{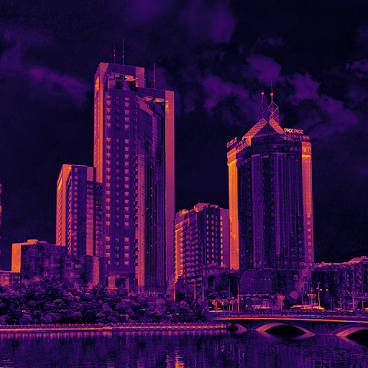}}\hfill
    \subfloat[CIEDE2000]{\includegraphics[width=0.19\linewidth]{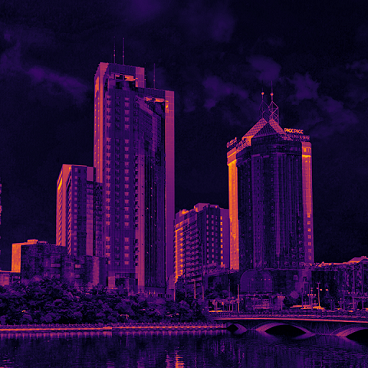}}\hfill
    \subfloat[S-CIELAB]{\includegraphics[width=0.19\linewidth]{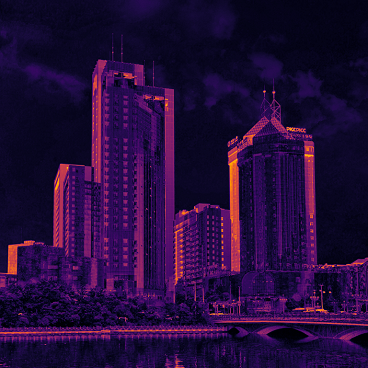}}\hfill
    \subfloat[Ouni08]{\includegraphics[width=0.19\linewidth]{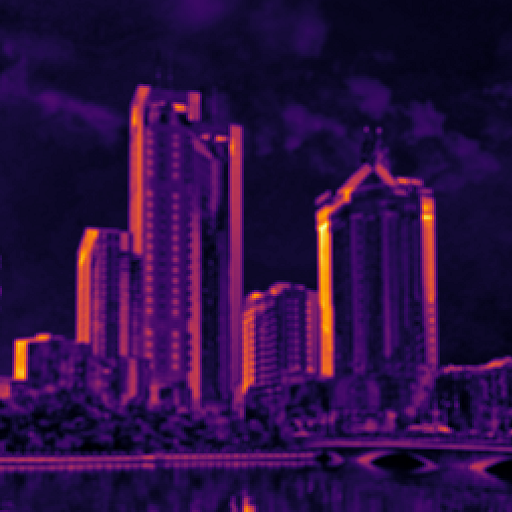}}\hfill
    \begin{subfigure}[t]{0.0084\linewidth}
        \includegraphics[width=0.0084\linewidth]{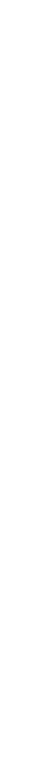}
    \end{subfigure}
    \\
    \subfloat[Image B]{\includegraphics[width=0.19\linewidth]{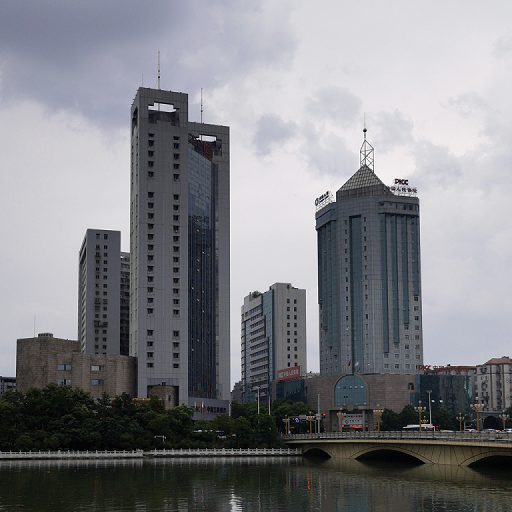}} \hfill
    \subfloat[\FLIP]{\includegraphics[width=0.19\linewidth]{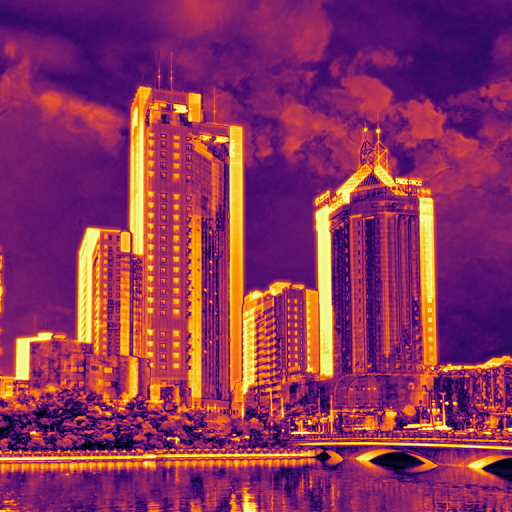}} \hfill
    \subfloat[CD-Net]{\includegraphics[width=0.19\linewidth]{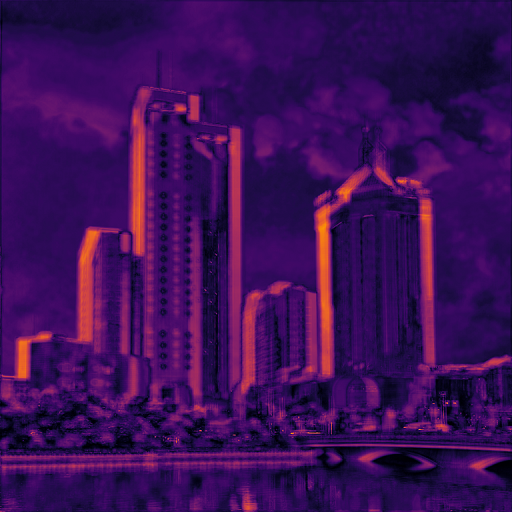}} \hfill
    \subfloat[CD-Flow]{\includegraphics[width=0.19\linewidth]{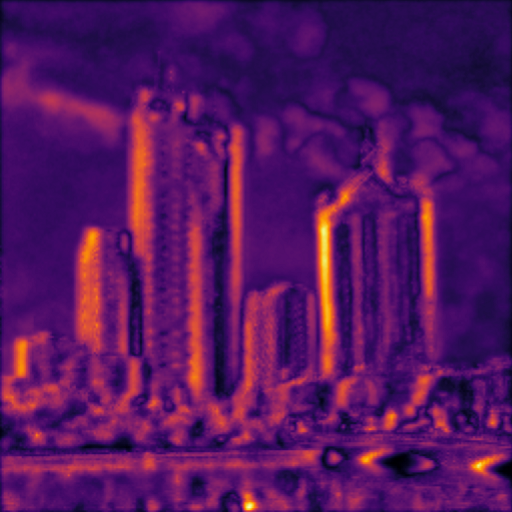}} \hfill
    \subfloat[MS-SWD]{\includegraphics[width=0.19\linewidth]{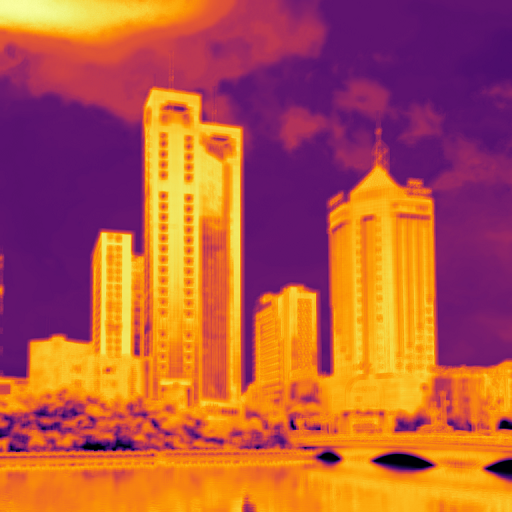}} \hfill
    \subfloat{\includegraphics[width=0.0084\linewidth]{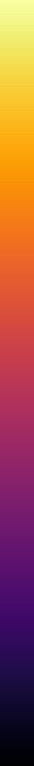}}
    \caption{Comparison of CD Maps for a non-perfectly aligned pair, where a warmer color indicates a larger pixel-wise (or patch-wise) CD.}
    \label{fig:cd_map}
\end{figure}

\noindent
\textbf{Visualization of CD Maps}.
We compare the CD maps generated by the proposed MS-SWD with seven other representative CD measures. Fig.~\ref{fig:cd_map} shows the visualization results for a non-perfectly aligned pair. It is evident that all competing methods are sensitive to image misalignment, leading to falsely large CDs along object boundaries. On the contrary, the proposed MS-SWD generates a more accurate CD map, correcting identifying areas of large CDs (\eg, the clouds, buildings, and trees).

\subsection{Ablation Studies}
\noindent
\textbf{Verification as an Empirical Metric}. We design computational experiments to verify that the proposed MS-SWD behaves empirically as a metric, which holds potential in perceptual optimization of color image processing algorithms. Non-negativity and symmetry are immediately apparent from Eq.~\eqref{eq:SWD_pyramid}. For the identity of indiscernibles (\ie, $\Delta E(\bm{X}, \bm{Y})=0 \Longleftrightarrow \bm{X}=\bm{Y}$), we resort to the reference image recovery task~\cite{ding2021comparison}  as a way of examining pixel-level image fidelity (see \cref{fig:img_recovery}), where we find MS-SWD successfully recovers the reference image from all structured and non-structured initializations. For the triangle inequality (\ie, $\Delta E(\bm{X}, \bm{Y}) \le \Delta E(\bm{X}, \bm{Z}) + \Delta E(\bm{Z}, \bm{Y})$), we test MS-SWD on $100,000$ randomly selected image triplets of the same content from SPCD, and find no violations. In conclusion, we empirically establish that MS-SWD behaves as a metric in the mathematical sense.

\noindent
\textbf{Number of Random Linear Unit Projections}. 
We investigate the effect of the number of random linear unit projections in MS-SWD, with $P$ values selected from $ \{4, 16, 64, 128, 256\}$. \Cref{tab:ablation_proj} shows the results on the non-perfectly aligned pairs from SPCD, where the average inference time is estimated using an NVIDIA A100 GPU. It is clear that the CD assessment performance of MS-SWD remains fairly stable when we decrease $P$, but excessively small $P$ values will compromise the ability of MS-SWD to maintain pixel-level image fidelity. Therefore, we choose $P=128$ to balance prediction accuracy, metric property, and computational complexity.

\begin{table}[t]
\caption{Ablation analysis of the number of random linear unit projections in MS-SWD. The default setting is highlighted in boldface.}
\label{tab:ablation_proj}
\centering
\begin{threeparttable} 
\begin{tabular}{c|ccc|c}
\toprule[1pt] 
$\#$ of random projections & STRESS$\downarrow$ & PLCC$\uparrow$ & SRCC$\uparrow$ & Time (ms)\\
\hline
\quad $P=4$ & $31.849$ & $0.804$ & $0.779$ & $3.7$  \\
\quad $P=16$ & $29.186$ & $0.833$ & $0.799$ & $4.2$  \\
\quad $P=64$ & $28.425$ & $0.841$ & $0.805$ & $6.2$  \\
\quad $P=\mathbf{128}$ & $28.363$ & $0.841$ & $0.805$ & $9.5$  \\
\quad $P=256$ & $28.318$ & $0.842$ & $0.806$ & $15.3$  \\
\bottomrule[1pt] 
\end{tabular}
\end{threeparttable}
\end{table}

\noindent\textbf{Learnable Non-Linear Projections}. To further enhance MS-SWD, we explore replacing random linear unit projections with learnable non-linear projections. Inspired by CD-Net~\cite{wang2022cdnet}, we design a lightweight neural network for nonlinear projection, including a front-end $11 \times 11$ convolution layer and a back-end $1\times 1$ convolution layer with leaky ReLU in between. Training involves minimizing the PLCC loss using Adam optimizer, initialized with a learning rate of $10^{-3}$ and decayed by a factor of $2$ every $5$ epochs. We train the network for $10$ epochs using a mini-batch size of $30$. We randomly partition SPCD into $70\%$, $10\%$, and $20\%$ for training, validation, and testing, respectively, while ensuring content independence during dataset splitting. This procedure is repeated ten times, and the average results are reported. As shown in \Cref{tab:ablation_train}, our learned MS-SWD outperforms the most advanced CD-Flow with just $0.08\%$ of its parameters.

\begin{table}[t]
\scriptsize
    \caption{Ablation analysis of using learnable non-linear projections in place of random linear unit projections in MS-SWD.}
    \label{tab:ablation_train}
    \centering
    \begin{threeparttable} 
    \begin{tabular}{l|ccc|ccc|ccc}
    \toprule[1pt]
    \multirow{2}*{Method}  & \multicolumn{3}{c|}{Perfectly aligned pairs} & \multicolumn{3}{c|}{Non-perfectly aligned pairs} & \multicolumn{3}{c}{All}\\
    \cline{2-10}
    & STRESS$\downarrow$ & PLCC$\uparrow$ & SRCC$\uparrow$ & STRESS$\downarrow$ & PLCC$\uparrow$ & SRCC$\uparrow$ & STRESS$\downarrow$ & PLCC$\uparrow$ & SRCC$\uparrow$\\
    \hline
    CD-Net~\cite{wang2022cdnet}&  $20.891$  & $0.867$ & $0.870$ & $22.543$ & $0.818$ & $0.776$& $21.431$ & $0.846$ & $0.842$\\
    CD-Flow~\cite{chen2023learning} & $\textbf{16.613}$ & $\textbf{0.896}$  & $\textbf{0.904}$  &  $\textbf{21.374}$  & $0.856$  &  $0.794$  & $\textbf{18.473}$  &  $0.871$  & $0.865$\\
    \hline
    MS-SWD (Learned) & $21.870$ & $0.894$ & $0.896$ & $22.359$ & $\textbf{0.876}$ & $\textbf{0.857}$ & $22.364$ & $\textbf{0.884}$ & $\textbf{0.889}$\\
    \bottomrule[1pt]
    \end{tabular}
    \begin{tablenotes}
    \item[] Trainable parameters: CD-Net ($0.01$M), CD-Flow ($60.49$M), and MS-SWD ($0.05$M).
    \end{tablenotes}
    \end{threeparttable}
\end{table}

\subsection{Image and Video Color Transfer}
\begin{figure}[t]
    \centering
    \subfloat[Target]{\includegraphics[width=0.32\linewidth]{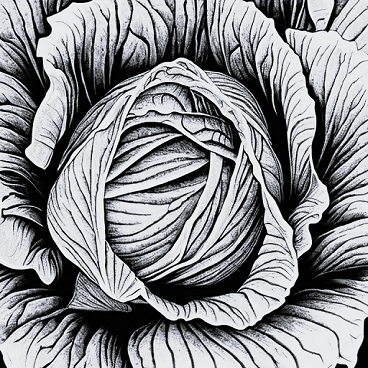}\label{fig:colorization1a}}\hspace{1.2mm}
    \subfloat[Source]{\includegraphics[width=0.32\linewidth]{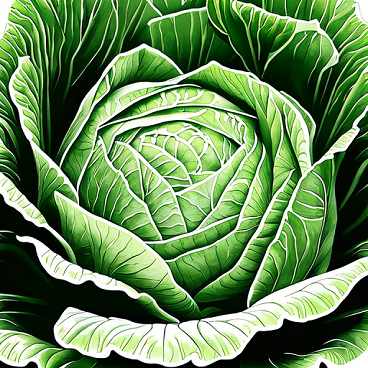}\label{fig:colorization1r}}\hspace{1.2mm}
    \subfloat[Output]{\includegraphics[width=0.32\linewidth]{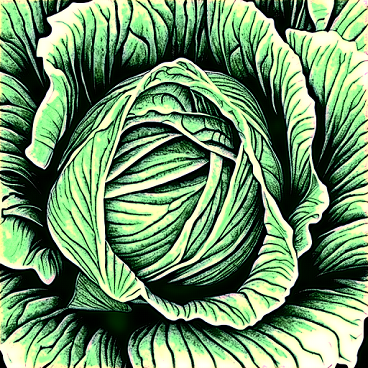}\label{fig:colorization1b}}
    \\
    \subfloat[Target]{\includegraphics[width=0.32\linewidth]{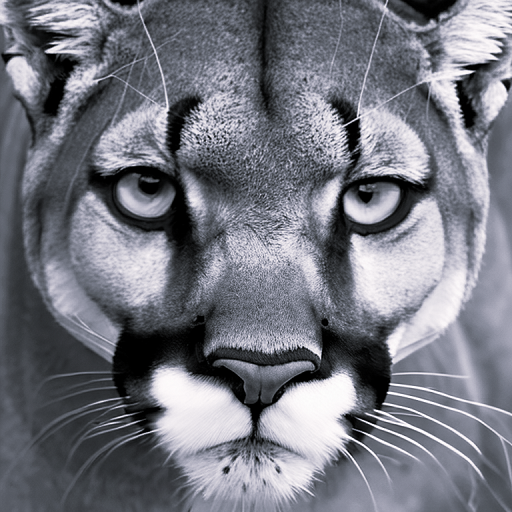}\label{fig:colorization2a}}\hspace{1.2mm}
    \subfloat[Source]{\includegraphics[width=0.32\linewidth]{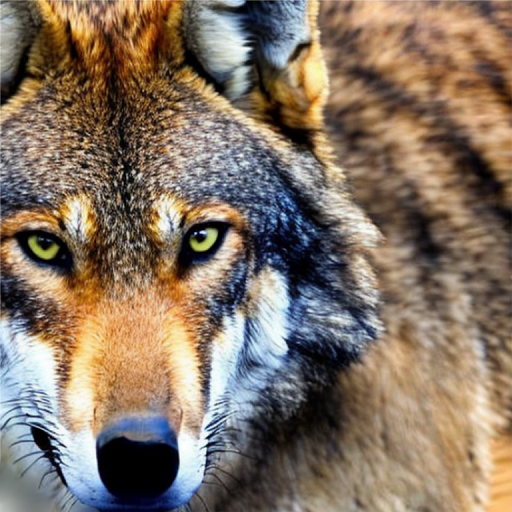}\label{fig:colorization2r}}\hspace{1.2mm}
    \subfloat[Output]{\includegraphics[width=0.32\linewidth]{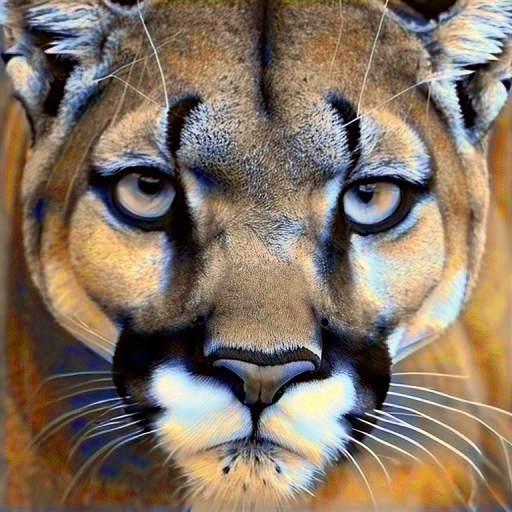}\label{fig:colorization2b}}
    \caption{Image color transfer results guided by MS-SWD.}
    \label{fig:colorization}
\end{figure}

\begin{figure*}[t]
  \centering
   \includegraphics[width=0.96\linewidth]{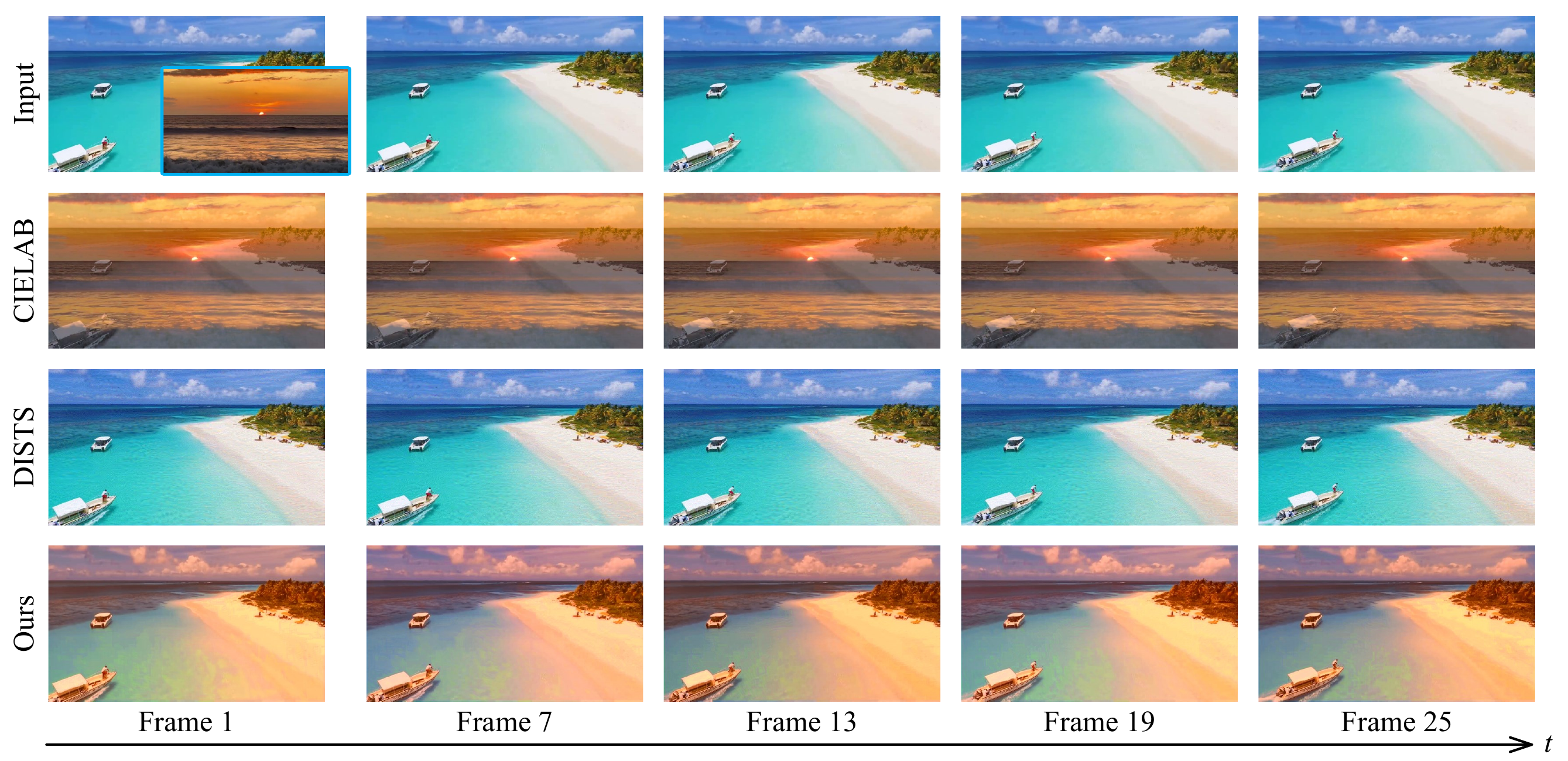}
   \caption{Comparison of video color transfer results. The first row displays five frames sampled from the target video, with the source image providing the desired color appearance shown in the bottom right corner of Frame 1.}
   \label{fig:color_transfer}
\end{figure*}

In this subsection, we explore the application of the proposed MS-SWD in the image and video color transfer task. Our computational algorithm is straightforward: given a source color image (or video) $\bm X$, we aim to transfer its color appearance to the target grayscale image (or color video)  $\bm Y_\mathrm{init}$ through the following optimization problem:
\begin{equation}
  \bm{Y}^\star=\arg \min_{\bm{Y}} \Delta E(\bm{X},\bm{Y}),
  \label{eq:color_transfer}
\end{equation}
starting from $\bm Y_\mathrm{init}$. The results of image color transfer are shown in \cref{fig:colorization}, demonstrating a successful color mapping from source to target, while preserving the underlying structure. The video color transfer outcomes are shown in \cref{fig:color_transfer}. Using CIELAB, we transfer the color patterns as well as unwanted structure details to the target. DISTS fails in this task, often leaving the target largely unchanged. MS-SWD produces visually appealing video frames in terms of transferred color appearance, structure preservation, and temporal consistency.

\section{Conclusion and Discussion}
\label{sec:conclusion}
We have introduced MS-SWD, the multiscale sliced Wasserstein distance designed for measuring perceptual CDs in photographic images. Unlike traditional \textit{co-located} comparisons prevalent in CD assessment, MS-SWD enables efficient comparisons between \textit{non-local} patches of similar structure and color information, making it exceptionally robust to real-world image misalignment. MS-SWD is training-free using random linear unit projections, which can be replaced by learnable non-linear projections for improved performance.

We highlight the importance of multiscale analysis in MS-SWD for preserving pixel-level image fidelity, thereby demonstrating its empirical metric properties. An intriguing mathematical inquiry remains: whether MS-SWD is indeed a metric, given specific hyper-parameter configurations. Additionally, exploring alternative linear and non-linear image pyramids beyond the Gaussian pyramid, such as the (normalized) Laplacian pyramid \cite{burt1983laplacian}, steerable pyramid \cite{simoncelli1995steerable_pyramid}, and VGG feature hierarchy \cite{simonyan2015vgg}, presents an interesting avenue. Last, there is potential to extend the non-local computation in MS-SWD (via the \texttt{sort}() operator) to measure other perceptual aspects of human vision (\eg, image quality).

\section*{Acknowledgements}
This work was supported in part by the National Key Research and Development Program of China (2023YFE0210700), the Hong Kong ITC Innovation and Technology Fund (9440390), the National Natural Science Foundation of China (62071407, 62375233, 62301323, 62441203, 62302423, and 62311530101), and the Shenzhen Natural Science Foundation (20231128191435002).

\bibliographystyle{splncs04}
\bibliography{main}

\end{sloppypar}
\end{document}